\documentclass[letterpaper, 10 pt, conference]{ieeeconf}  

\IEEEoverridecommandlockouts                              

\usepackage{graphicx}
\usepackage{xcolor}
\usepackage{bm}
\usepackage{multirow}
\usepackage{booktabs}
\usepackage{tabularx}
\usepackage{url}
\usepackage{amsmath} 
\usepackage{amssymb}  
\usepackage[normalem]{ulem}
\useunder{\uline}{\ul}{}

\title{\LARGE \bf
PCDepth: Pattern-based Complementary Learning for Monocular Depth Estimation by Best of Both Worlds
}
\author{Haotian Liu\textsuperscript{1} \qquad Sanqing Qu\textsuperscript{1} \qquad Fan Lu\textsuperscript{1} \qquad Zongtao Bu\textsuperscript{1} \\
\qquad Florian Röhrbein\textsuperscript{2} \qquad Alois Knoll\textsuperscript{3} \qquad Guang Chen\textsuperscript{1*}\thanks{*Corresponding author: guangchen@tongii.edu.cn} \\
{\small \textsuperscript{1}Tongji University\qquad 
\textsuperscript{2}Chemnitz University of Technology\qquad 
\textsuperscript{3}Technical University of Munich}}

\begin{document}

\maketitle
\thispagestyle{empty}
\pagestyle{empty}

\begin{abstract}
  Event cameras can record scene dynamics with high temporal resolution, providing rich scene details for monocular depth estimation (MDE) even at low-level illumination. Therefore, existing complementary learning approaches for MDE fuse intensity information from images and scene details from event data for better scene understanding. 
  However, most methods directly fuse two modalities at pixel level, ignoring that the attractive complementarity mainly impacts high-level patterns that only occupy a few pixels. For example, event data is likely to complement contours of scene objects.
  In this paper, we discretize the scene into a set of high-level patterns to explore the complementarity and propose a Pattern-based Complementary learning architecture for monocular Depth estimation (PCDepth). 
  Concretely, PCDepth comprises two primary components: a complementary visual representation learning module for discretizing the scene into high-level patterns and integrating complementary patterns across modalities and a refined depth estimator aimed at scene reconstruction and depth prediction while maintaining an efficiency-accuracy balance. Through pattern-based complementary learning, PCDepth fully exploits two modalities and achieves more accurate predictions than existing methods, especially in challenging nighttime scenarios. Extensive experiments on MVSEC and DSEC datasets verify the effectiveness and superiority of our PCDepth. Remarkably, compared with state-of-the-art, PCDepth achieves a 37.9\%  improvement in accuracy in MVSEC nighttime scenarios.

\end{abstract}

\begin{figure}[t]
    \centering
    \includegraphics[width=0.95\linewidth]{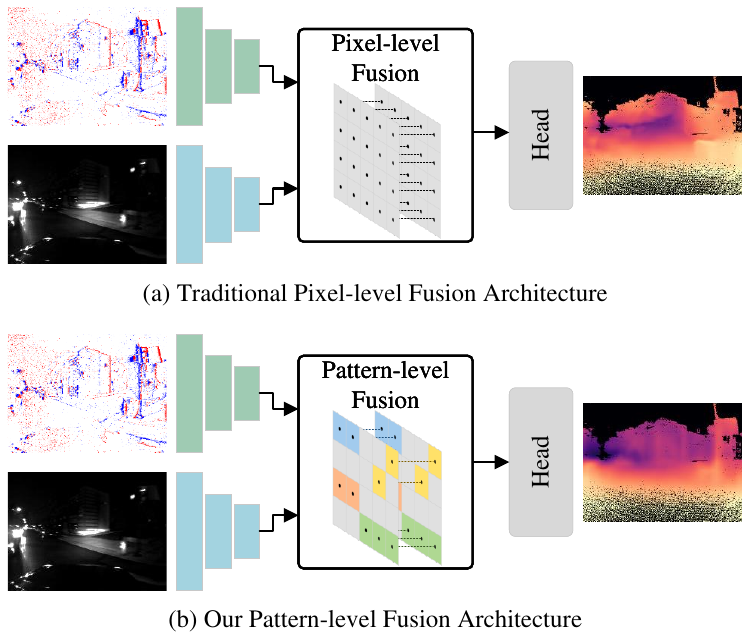}
    \vspace{-0.1in}
    \caption{Existing pixel-level fusion architecture \emph{vs} our pattern-level fusion architecture. (a) Mainstream works take direct pixel-level fusion of features of two modalities, ignoring that the complementarity mainly impacts patterns of scene objects, which only occupy a few pixels. (b) Instead, our proposed pattern-level fusion architecture emphasizes the importance of high-level patterns of scene objects and conducts complementary learning of visual patterns, which achieves high-quality depth estimates.} 
    \label{fig:patternvspixel}
\end{figure}

\section{INTRODUCTION}

Monocular depth estimation (MDE) aims to compute the distance between the projection center and the 3D point corresponding to each pixel and is a basic topic in event-based vision~\cite{survey, surveychen, surveydeep}. It plays a role in many applications, such as ego-motion estimation~\cite{zhuego, yeego} and visual odometry~\cite{eaodo}. Different from images that capture dense intensity, event data only records scene dynamics. Namely, event cameras output no recordings in static scenes (regardless of noise), leading to limited depth estimation. Therefore, several works~\cite{ramnet, even, srfnet} take advantage of both modalities for better prediction accuracy. Through well-designed fusion modules, these methods aggregate global intensity and local details at pixel level. Despite decent progress, these methods ignore that the complementarity mainly impacts high-level patterns instead of all pixels. For example, event data records contour information of scene objects solely and per-pixel intensity from images at low-level illumination is not fully reliable. 

To this end, a natural thought is to take the best of both worlds at pattern level rather than pixel level. In a recent work~\cite{idisc}, researchers succeed in learning a finite set of visual patterns to represent the whole scene and achieve encouraging performance for MDE. Building upon this, we propose to delve deeper into the complementarity at pattern level and implement a pattern-level fusion of event and image modalities in this paper. Fig.~\ref{fig:patternvspixel} illustrates the difference between existing pixel-level fusion architectures and our proposed pattern-level fusion architecture.

To materialize our idea, we propose a Pattern-based Complementary learning architecture for monocular Depth estimation (PCDepth). Technically, we revamp the dominant visual pattern learning framework with two different components. We first propose a complementary visual representation learning module to generate a unified set of tokens representing the scene of both modalities. By discretizing dual scenes into two sets of visual tokens with transposed attention and integrating complementary patterns through a score-based fusion module, scene information is consolidated. Then, we delicately design a depth estimator to reconstruct the scene and derive predictions. To maintain an accuracy-efficiency balance, scene reconstruction is achieved by cross-attention limited to a single spatial resolution and depth estimates are fine-tuned using GRU blocks. Thanks to the pattern-based complementary learning, PCDepth adaptively gathers content from images and details from events and thus improves prediction quality, especially in challenging nighttime scenarios.

We evaluate PCDepth on DSEC~\cite{dsec} and MVSEC~\cite{mvsec} datasets. Extensive experiments demonstrate the superiority.

In summary, our main contributions are as follows:
\begin{itemize}
    \item We argue that the complementarity between event and image modalities mainly impacts high-level patterns and propose a pattern-based complementary learning framework to explore the attractive complementarity.
    \item We propose PCDepth, which comprises two main components: a complementary visual representation learning module and a refined depth estimator.
    \item Thanks to pattern-based complementary learning, PCDepth gathers both modalities and achieves high accuracy, especially in nighttime scenarios. Compared with state-of-the-art, PCDepth achieves a 37.9\% accuracy improvement in MVSEC nighttime scenarios.
\end{itemize}

\section{Related Work}
\textbf{Image-based learning.} Recently, learning-based methods~\cite{eigen, fcrn, adabins, crf, dpt} dominate the MDE field. Acting as a milestone, Eigen et al.~\cite{eigen} first introduce a coarse-to-fine convolutional network to learn monocular depth. Since then, a large number of novel learning-based architectures merge. FCRN~\cite{fcrn} proposes an encoder-decoder network with fully residual blocks and learnable upsampling convolutions. DORN~\cite{dorn} recasts the MDE as an ordinal regression problem by discretizing depth as a spacing-increasing distribution and training the network with an ordinary regression loss, handling MDE from a new respective. 

More recently, Vision Transformer (ViT)~\cite{transformer} shows great potential in dense tasks. Bins series~\cite{adabins, localbins, binsformer, iebins} extend the pipeline of depth discretization and combination and exploit Transformer blocks to better process global information of intermediate features. DPT~\cite{dpt} proposes a unified Transformer architecture for dense prediction, which can be applied to monocular depth and semantic segmentation. NeWCRF~\cite{crf} combines CRFs~\cite{crf1} and multi-head attention~\cite{attention} to optimize depth estimates sparse-to-densely. IDisc~\cite{idisc} discretizes the scene into visual tokens and projects them back to the pixel plane for MDE with attention, realizing depth prediction in a continuous-discrete-continuous manner.

We draw inspiration from IDisc to design our complementary learning framework. In contrast to IDisc which learns high-level patterns from image modality solely, we extend the pattern-based representation learning to explore the complementarity between event and image modalities.

\textbf{Event-based Learning.} Inspired by the success in the image domain, existing event-based methods for MDE transform event data into grid-like representations to apply learning-based architectures~\cite{e2depth, erformer, ofdepth, hmnet}. E2Depth~\cite{e2depth} incorporates a recurrent convolutional block into UNet~\cite{unet} to aggregate the temporal consistency of event data to improve predictions. Mixed-EF2DNet~\cite{ofdepth} incorporates an optical flow network to align event data, which compensates for the lost temporal information and improves predictions. EReFormer~\cite{erformer} proposes a pure transformer architecture with a recursive mechanism, considering global context and temporal information simultaneously. HMNet~\cite{hmnet} encodes event contents at an adaptive scale, taking short-term and long-term dependencies into account.

\textbf{Complementary learning of event and RGB modalities.} Due to the inherent spatial sparsity of event data and lack of intensity, one single event modality cannot adapt to all circumstances. Therefore, many learning-based approaches~\cite{ramnet, srfnet, even} take the best of both worlds to improve the scene understanding. RAMNet~\cite{ramnet} extends E2Depth by exploiting the intensity from the image domain. EVEN~\cite{even} proposes a low-light enhancement module to enhance image quality and fuse both modalities, which is suitable for adverse night conditions. SRFNet~\cite{srfnet} learns a spatial reliability mask from both modalities to guide the inter-modal feature fusion and depth refinements. 

Contrary to mainstream works that design delicate fusion modules to aggregate features of different modalities at pixel level, we instead emphasize that complementarity mainly impacts high-level patterns and propose a pattern-based complementary learning framework to boost accuracy.

\section{Methodology}
Given an image $\bm{I}$ and corresponding event stream $\{\mathcal{E}\}$, MDE aims to estimate a dense map $\bm{\mathrm{d}}$ representing the per-pixel depth of 3D scenes. MDE is an ill-posed problem because each pixel can correspond to infinite 3D points. Mainstream methods~\cite{ramnet, srfnet, even} take advantage of learning architectures to encode two modalities and impose delicate per-pixel fusion on feature maps. However, the complementarity between event and image modalities mainly impacts high-level patterns of the scene, which only occupy a small percentage of pixels. To realize a more concise fusion, we discretize scenes into visual tokens and aggregate complementary patterns. The visual tokens combine object contents and contour details from two modalities, yielding high-quality depth predictions.

In the following, we first describe the data preparation and then present the detailed architecture of PCDepth.

\begin{figure*}[t]
    \centering
    \includegraphics[width=0.97\linewidth]{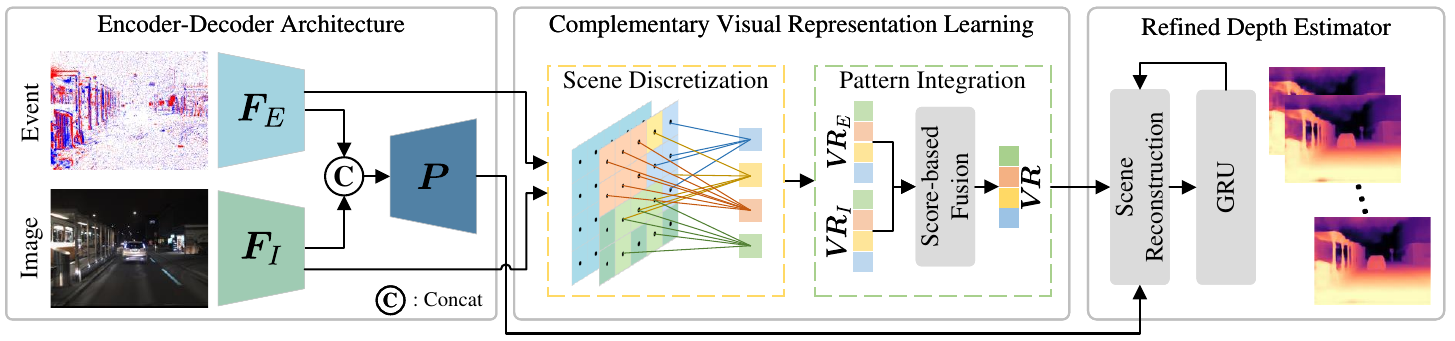}
    \vspace{-0.1in}
    \caption{The overall architecture of PCDepth. First, we use a general encoder-decoder pipeline to process event and image inputs. Encoded features from event and image modalities are derived from dual feature extractors. These two features are concatenated and sent to decoder layers to generate pixel embeddings. Second, the proposed complementary visual representation learning module discretizes features of both modalities through transposed attention and integrates complementary patterns into one set of visual tokens through score-based fusion. Finally, we reconstruct the scene to one single scale through cross-attention and exploit GRU blocks to refine depth estimates, maintaining an accuracy-efficiency balance. } 
    \vspace{-0.2in}
    \label{fig:overall}
\end{figure*}
\subsection{Data preparation}
Each event in the event stream $\{\mathcal{E}\}$ is a triplet $\bm{e}_k = (\bm{x}_k, p_k, t_k)$ containing triggered location $\bm{x}_k = (x_k,y_k)^T$, timestamp $t_k$ and polarity $(p_k=\pm 1)$. To ease the feature encoding with learning-based architectures, we transform the event stream into a 3D volume representation $\bm{E}(x, y, t)$ along the time axis following~\cite{zhuego, e2depth, ramnet}. $\bm{E}(x, y, t)$ is generated as:
\begin{align}
&t_{i}^{*} =(B-1)(t_{i}-t_{1}) /(t_{N_e}-t_{1}) \\
&\bm{E}(x, y, t) =\sum_{i} p_{i} k_{b}(x-x_{i}) k_{b}(y-y_{i}) k_{b}(t-t_{i}^{*}) \\
&k_{b}(a) =\max (0,1-|a|),
\end{align}
where $t_{i}^{*}$ represents the $i$-th time bin. $B$ and $N_e$ represent the length of time bins and the number of events, respectively. $k_{b}(a)$ refers to a bilinear interpolation function. Note that other event representations~\cite{tore, stereorep, learnrep} are also possibly suitable for MDE, but the evaluation of different representations is not the core of this work.  

\subsection{PCDepth Architecture}
PCDepth adopts the basic encoder-decoder architecture in IDisc~\cite{idisc}. The event representation and image are separately entered into two feature extractors to achieve multi-scale encoded features. Encoded features of both modalities are then concatenated and fed into a decoder to generate multi-scale pixel embeddings. On top of the basic pipeline, one core component, the complementary visual representation learning module is proposed to explore pattern-based complementarity. The complementary visual representation learning module discretizes encoded features and integrates complementary patterns into one unified set of visual tokens. Besides, a refined depth estimator is designed to reconstruct the scene limited to a single scale and refine depth estimates with GRU blocks, maintaining an accuracy-efficiency balance. 
Fig.~\ref{fig:overall} summarizes the architecture of PCDepth.

\textbf{Complementary Visual Representation Learning}. Instead of fusing features directly at pixel level, which is a common technique in mainstream works~\cite{ramnet, even, srfnet}, we believe that images and events complement each other at the pattern level, allowing us to leverage their complementary patterns. To introduce it into our framework, we first discretize encoded features of both modalities into two sets of visual tokens respectively and then integrate complementary patterns through a score-based fusion module.

\begin{figure}[t]
    \centering
    \includegraphics[width=0.8\linewidth]{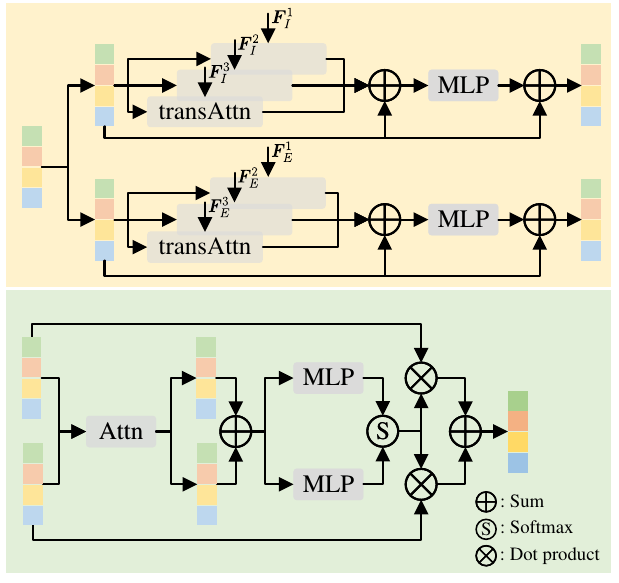}
    \caption{Complementary visual representation learning. (Top) We use transposed attention to discretize both modalities into two sets of visual tokens. (Bottom) Each set of visual tokens is first enhanced through standard attention. Then the sum of enhanced visual tokens is entered into MLPs and Softmax operator to generate two score maps for both modalities.} 
    \vspace{-0.2in}
    \label{fig:fuse}
\end{figure}

Following existing works~\cite{idisc, slotattention}, we use iterative transposed attention for scene discretization. Meanwhile, considering that multi-scale encoded features contain information at different levels, we extend the transposed attention to discretize multi-scale features simultaneously. 
The top of Fig.~\ref{fig:fuse} shows the overview of scene discretization. In detail, given multi-scale encoded features $\bm{F}_E^i, \bm{F}_I^i, i=1,2,3$ from event modality and image modality, we first initialize a set of visual tokens $\bm{V\!R}\in \mathbb{R}^{N\times C}$ following a normal distribution, where $N$ and $C$ mean the number and dimension. The $k$-th iteration of scene discretization is achieved by:
\begin{align}
    &\bm{V\!R}^k = \bm{V\!R}^{k-1} + \frac{1}{3} [\sum_{i=1}^{3}\operatorname{transAttn}(\bm{V\!R}^{k-1}, \bm{F}^i)],  \\
    &\bm{V\!R}^k = \bm{V\!R}^k + \operatorname{MLP}(\bm{V\!R}^k),
\end{align}
where $\operatorname{transAttn}(\bm{A}, \bm{B})$ represents transposed attention operation and query, key and value are generated from linear projections as $f_{\bm {Q}}(\bm{A})$, $f_{\bm{K}}(\bm{B})$, $f_{\bm {V}}(\bm{B})$. We drop the modality superscript of the encoded features since we employ the same operation to get $\bm{V\!R}_E^k$ and $\bm{V\!R}_I^k$ for both modalities.

After obtaining $\bm{V\!R}_E^k$ and $\bm{V\!R}_I^k$, the following routine procedure is to add or concatenate them to aggregate complementary patterns. However, we notice that the complementarity between event and image is not entirely equivalent. Images demonstrate superiority in daytime conditions with adequate illumination, whereas event data exhibits better adaptability in low-light environments. Direct concatenation or summation could potentially introduce incorrect scene information. To this end, we balance two modalities through a learnable score map. The bottom of Fig.~\ref{fig:fuse} displays the score-based fusion module for pattern aggregation. Technically, $\bm{V\!R}_E^k$ and $\bm{V\!R}_I^k$ are enhanced through cross-attention to emphasize consistent patterns. Then the sum of enhanced visual tokens is entered into MLPs and Softmax to generate score maps $\bm {s}_I^k$ and $\bm {s}_E^k$. The complementary visual token $\bm{V\!R}$ at $k$-th iteration is achieved as: 
\begin{align}
    \bm{V\!R}^k = \bm {s}_I^k \cdot \bm{V\!R}_I^{k} + \bm {s}_E^k \cdot \bm{V\!R}_E^{k}.
\end{align}


\begin{figure}[t]
    \centering
    \includegraphics[width=0.85\linewidth]{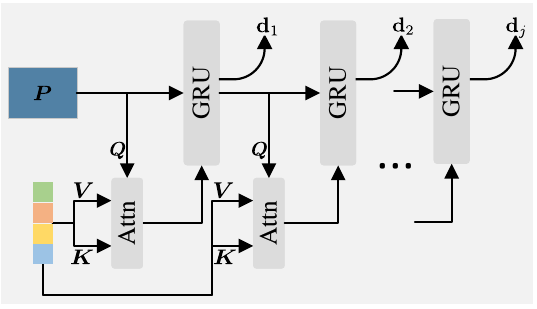}
    \caption{The refined depth estimator design. To maintain an accuracy-efficiency balance, we transform the multi-scale pixel embeddings into one scale and exploit GRU to refine the depth estimates.} 
    \label{fig:LDE}
\end{figure}

\textbf{Refined Depth Estimator.} Recall that MDE requires per-pixel values, whereas pattern-level complementary learning generates visual tokens representing discretized scene objects. Hence, scene reconstruction is necessary before depth prediction. A 
standard practice is projecting the visual tokens back onto the pixel plane through cross-attention. To circumvent heavy computational burdens resulting from cross-attention on multi-scale pixel embeddings, we opt for one low prediction scale and utilize GRU iterations for depth refinement, contributing to the accuracy-efficiency trade-off.

Technically, given multi-scale pixel embeddings $\bm{P}^i, i=1,2,3$, we first transform them into a fixed scale $\bm{P_0}$ with PixelUnshuffle~\cite{pixelshuffle} as initial hidden state. At each GRU stage of $G_2$ iterations, we project back visual tokens through cross-attention and feed them into the GRU to update the hidden state. The $j$-th iteration is obtained as:
\begin{align}
    \bm{P}_j &= \operatorname{Attn}(\bm{H}_{j-1}, \bm{V\!R}),\\
    \bm{H}_j &= \operatorname{GRU}(\bm{H}_{j-1}, \bm{P}_j),
\end{align}
where $\operatorname{Attn}$ represents the cross-attention layer, $\bm{H}_j$ represents the $j$-th hidden state of the GRU block. Especially, the initial hidden state $\bm{H}_0$ is generated by a $\operatorname{tanh}$ activation of $\bm{P}_0$. At each stage of iterations, the depth is predicted by $\bm{H}_j$ through a simple convolutional block and upsampled to the original scale through convex upsampling~\cite{raft, practicalstereo}.

\textbf{Post-Process and Supervison.} Following RAMNet~\cite{ramnet}, our framework first outputs a normalized log depth $\hat {\bm{\mathrm{d}}} \in [0, 1]$, which facilitates depth prediction for diverse scenes. Scaled depth $\bm{\mathrm{d}}$ is then recovered as:
\begin{equation}
    \bm{\mathrm{d}} = \bm{\mathrm{d}}_{max} \operatorname{exp}(\frac{\bm{\mathrm{d}}_{max}}{\bm{\mathrm{d}}_{min}}(\hat {\bm{\mathrm{d}}}-1)),
\end{equation}
where $\bm{\mathrm{d}}_{max}$ and $\bm{\mathrm{d}}_{min}$ are prior maximum and minimum depth values for different datasets.
We use the well-established $\operatorname{SI_{log}}$ loss~\cite{idisc, eigen} to supervise our model. The loss function $\mathcal{L}$ is formally defined as: 
\begin{align}
    \mathcal{L} &= \sum_{j=1}^{G_2} \gamma^{G_2-j}\alpha\sqrt{\mathbb{V}(\delta_j)+\lambda\mathbb{E}(\delta_j)},\\
    \delta_j&= \operatorname{log}(\bm{\mathrm{d}}_j) -\operatorname{log}(\bm{\mathrm{d}}_{gt}),
\end{align}
where $\bm{\mathrm{d}}_j$ is the $j$-th prediction and $\bm{\mathrm{d}}_{gt}$ is the ground-truth. $\mathbb{V}(\delta)$ and $\mathbb{E}(\delta)$ compute empirical
variance and expected value of $\delta$. The two weights $\alpha$ and $\lambda$ for $\operatorname{SI_{log}}$ loss are set to 10 and 0.15. The weight $\gamma$ to balance different stages of predictions is set to 0.8.

\begin{table*}[t]
\centering
\caption{Comparison of performances on MVSEC dataset. $\uparrow$ denotes higher is better and $\downarrow$ denotes lower is better. G means the iteration of scene discretization and reconstruction. [S] means the method is pre-trained on synthetic datasets. The best results are highlighted in bold and the second-best results are underlined.}
\label{table:mvsec}
\resizebox{\textwidth}{!}{%
\begin{tabular}{lc|cccccc|cccccc}
\toprule
\multirow{2}{*}{} &
  \multirow{2}{*}{Input} &
  \multicolumn{6}{c|}{outdoor day1} &
  \multicolumn{6}{c}{outdoor   night1} \\
  & &
  $a1\uparrow$ & $a2\uparrow$ & $a3\uparrow$ & REL$\downarrow$ & RMS$\downarrow$ & RMSlog$\downarrow$ &
  $a1\uparrow$ & $a2\uparrow$ & $a3\uparrow$ & REL$\downarrow$ & RMS$\downarrow$ & RMSlog$\downarrow$ \\ \midrule
E2Depth~\cite{e2depth}      & E     & 0.567     & 0.772     & 0.876     & 0.346     & 8.564     & 0.421     & 0.408     & 0.615     & 0.754     & 0.591     & 11.210     & 0.646             \\
\midrule
RAMNet~\cite{ramnet}         & I+E & 0.541     & 0.778     & 0.877     & 0.303     & 8.526     & 0.424     & 0.296     & 0.502     & 0.635     & 0.583     & 13.340     & 0.830           \\ 
SRFNet~\cite{srfnet}         &I+E  &     0.637      &  0.810         &    0.900       &     0.268       &       8.453    &    0.375       &   0.433        &       0.662    &      0.800     &  0.371         &     11.469      &     0.521           \\
HMNet~\cite{hmnet}  &I+E &    \textbf{0.717} &\textbf{0.868} & {\ul0.940} &{\ul0.230} &6.922 &{\ul0.310} &0.497 &0.661 &0.784 &0.349 &10.818 &0.543\\
PCDepth[S] & I+E  &  0.688	&0.859	&0.936& {\ul 0.230}	&6.964	&0.318 &{\ul0.504}	&0.698	&0.824
 &{\ul0.313}	&10.035	&0.497 \\
 \midrule
Baseline-G4 &I+E& 0.697&	0.856&	0.937& 0.243&	{\ul6.667}	&0.312& \textbf{0.632}	&{\ul0.818}&	{\ul0.918}& \textbf{0.271}	&{\ul6.863}	&{\ul0.357}\\
PCDepth   &I+E    &	{\ul0.712} &	{\ul0.867} &	\textbf{0.941}	&\textbf{0.228}	&\textbf{6.526}&	\textbf{0.301 }&	\textbf{0.632}&	\textbf{0.822}	&\textbf{0.922}  &\textbf{0.271}	&\textbf{6.715}	&\textbf{0.354}  \\ 
\bottomrule

\end{tabular}%
}
\end{table*}

\begin{figure*}[t]
    \centering
    \includegraphics[width=0.97\linewidth]{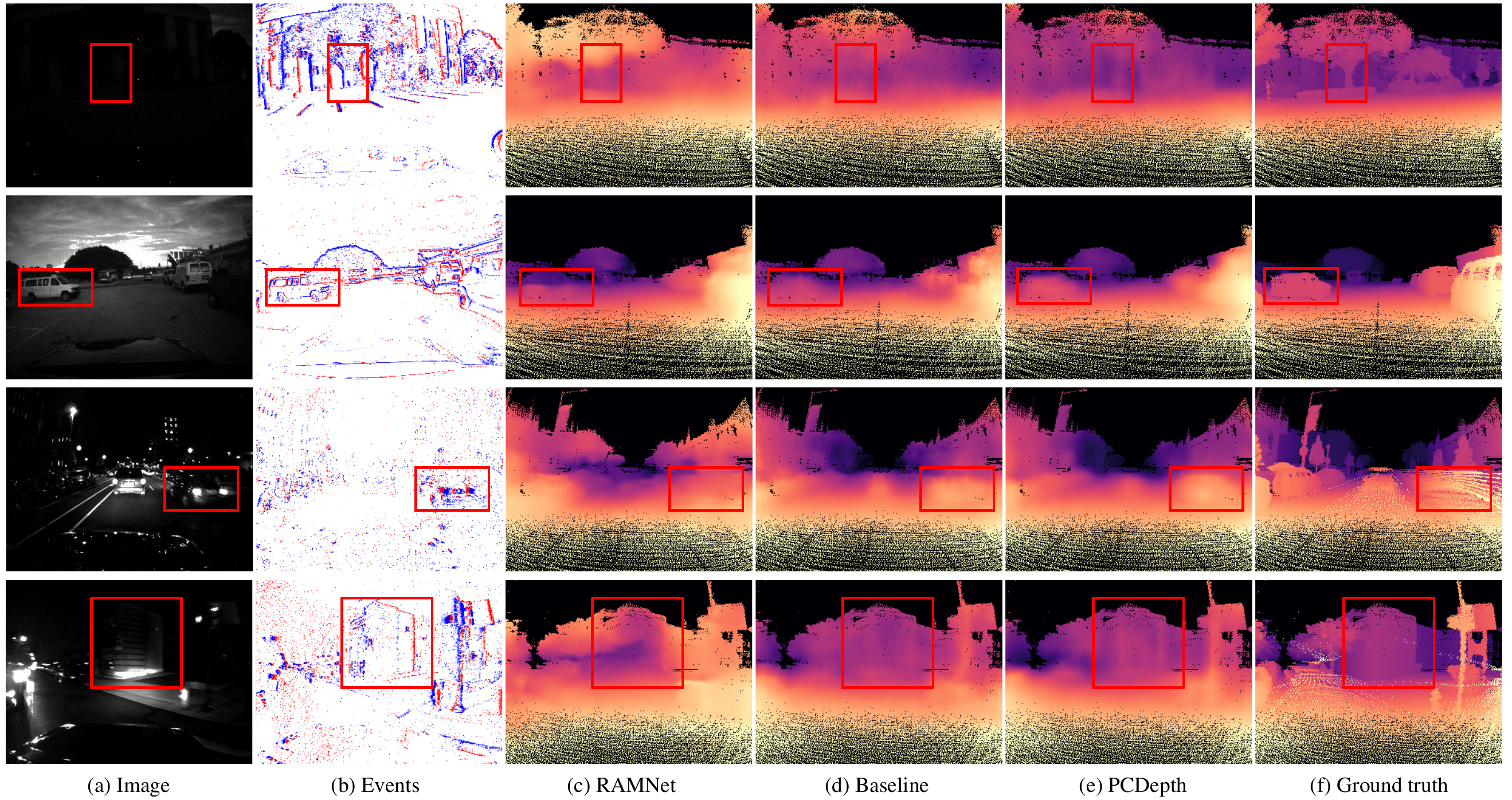}
    \caption{Qualitative results of MVSEC. We show the reference image and event frame in (a) and (b). (c) - (f) show results from RAMNet, baseline, our PCDepth and ground truth, respectively. Improvements are highlighted by red boxes.} 
    \label{fig:qua_mvsec}
\end{figure*}

\begin{table*}[t]
\centering
\caption{Results on DSEC dataset. $\uparrow$ denotes higher is better and $\downarrow$ denotes lower is better. G means the iteration of scene discretization and reconstruction. We adjust iterations of the baseline and our PCDepth for a comprehensive comparison. The best results are highlighted in bold.}
\label{table:dsec}
\resizebox{\textwidth}{!}{%
\begin{tabular}{lc|cccccc|cccccc}
\toprule
\multirow{2}{*}{} &
  \multirow{2}{*}{Input} &
  \multicolumn{6}{c|}{daytime sequences} &
  \multicolumn{6}{c}{nighttime sequences} \\
  & &
  $a1\uparrow$ & $a2\uparrow$ & $a 3\uparrow$ & REL$\downarrow$ & RMS$\downarrow$ & RMSlog$\downarrow$ &
  $a1\uparrow$ & $a2\uparrow$ & $a3\uparrow$ & REL$\downarrow$ & RMS$\downarrow$ & RMSlog$\downarrow$ \\ \midrule

IDisc~\cite{idisc} & I     &  0.873 &  0.970  &  0.993   & 0.106   &   3.973   & 0.155    &0.872 & 0.970&0.990 & 0.115    &4.266  &    0.165    \\
Baseline-G2   & I+E &  0.893    &0.979   &  0.996   &  0.099   &   3.517  &   0.139   &  0.884    &0.978    & 0.993    &  0.110   &  4.002    & 0.152  \\
PCDepth-G2 &  I+E     &    0.912  &0.984      & \textbf{0.997}   & \textbf{0.089}  & 3.349     & 0.127    &    0.891  &  0.981   &  0.995  & 0.105    & 3.860 &  0.145   \\
Baseline-G4 & I+E & 0.891 & 0.978 & 0.996 & 0.099 & 3.544 & 0.137 & 0.886 & 0.979 & 0.994 & 0.109 & 3.974 & 0.150\\
PCDepth  & I+E       &   \textbf{0.913}&  \textbf{0.985}  &  \textbf{0.997} & \textbf{0.089}  &   \textbf{3.311}   &  \textbf{0.126} &   \textbf{0.895}  & \textbf{0.983}  & \textbf{0.996}& \textbf{0.104} & \textbf{3.831}    &\textbf{0.144}      \\\bottomrule

\end{tabular}%
}
\end{table*}

\begin{figure*}[t]
    \centering
    \includegraphics[width=0.97\linewidth]{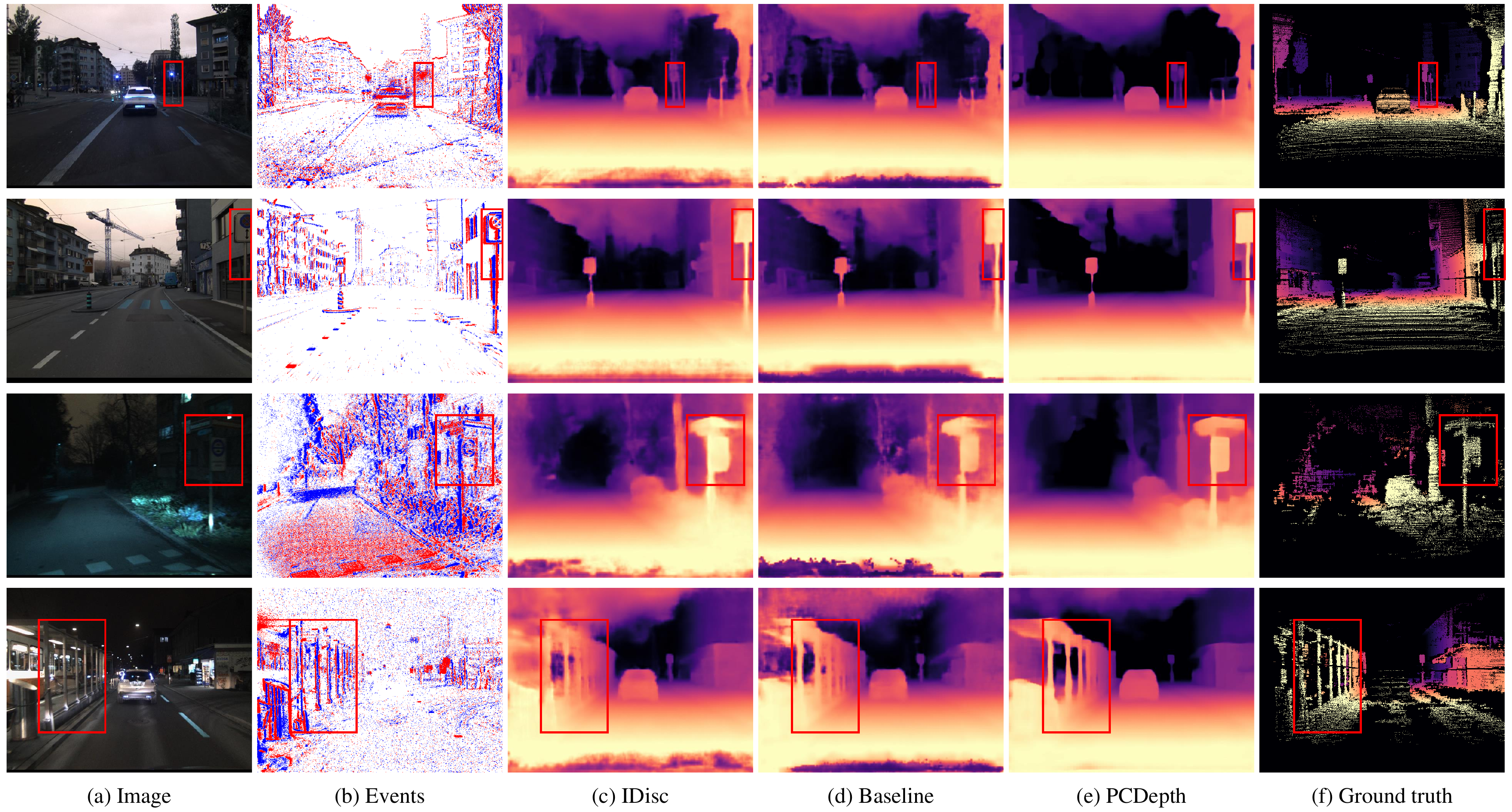}
    \caption{Qualitative results of DSEC. We show the reference image and event frame in (a) and (b). (c) - (f) show results from IDisc, baseline, our PCDepth and ground truth, respectively. Improvements are highlighted by red boxes.} 
    \label{fig:qua_dsec}
\end{figure*}

\begin{figure}[t]
    \centering
    \includegraphics[width=0.97\linewidth]{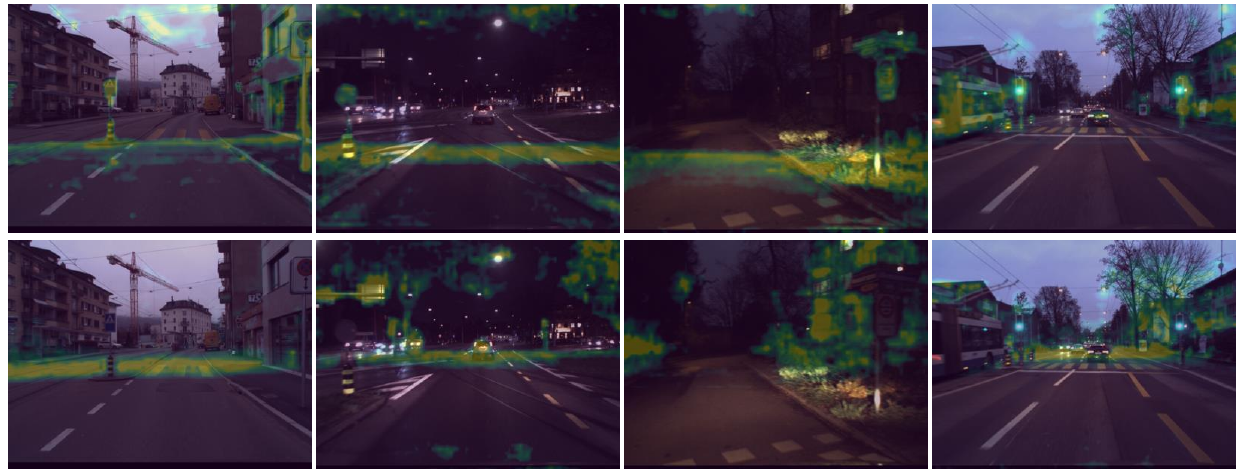}
    \caption{Attention maps on DSEC dataset. Each column presents attention maps of two different visual representations for one test image. } 
    \label{fig:attn_maps}
\end{figure}

\section{Experiments}
\textbf{Datasets and evaluation setup.} 
We provide extensive comparison results on two event-based datasets MVSEC~\cite{mvsec} and DSEC~\cite{dsec}, which both contain daytime and nighttime scenarios. For MVSEC, we follow the experimental setting in previous works~\cite{ramnet, srfnet} and train methods on synthetic EventScape dataset and outdoor\_day2 and evaluate outdoor\_day1 and outdoor\_night1. Besides, taking the large gap between daytime and nighttime scenarios into account, we incorporate outdoor\_night2 and outdoor\_night3 sequences as training data instead of pretraining models on EventScape to boost the performance. For DSEC, we convert disparity ground truth to depth\footnote{DSEC benchmark provides guidance for depth generation from disparity. Please refer to \url{https://dsec.ifi.uzh.ch/data-format/}.} since DSEC is not specifically designed for depth estimation. We manually split the training data into train and test subsets, maintaining the daytime-nighttime ratio as stereo evaluation.

\textbf{Metrics.}
As customary, we utilize absolute relative error (REL), root mean squared error (RMS) and its log variant (RMSlog) and the percentage of inlier pixels ($ai$) with thresholds $1.25^i$ for comprehensive evaluation.

\textbf{Implementation details.} PCDepth is implemented by PyTorch. We use events of 20ms for DSEC and events of 50ms for MVSEC, since events in MVSEC are much sparser. The channel of event representation $B$ is set to 3. We set two iteration numbers $G_1=G_2=4$. The number and length of visual representation is set to 32 and 128. The decoder is identical to IDisc's model. In training, we exploit AdamW~\cite{adamw} optimizer and One-cycle learning rate scheduler~\cite{onecycle}. For DSEC, We train models for 60K steps with a learning rate of 0.0002 and a batch size of 6. For MVSEC, we train models for 25K steps with a learning rate of 0.00001 and a batch size of 8.

\begin{table}[t]
\centering
\caption{Pattern Fusion Ablations.}
\label{table:ablation_fuse}
\addtolength{\tabcolsep}{1.0pt}
\resizebox{0.45\textwidth}{!}{
    \begin{tabular}{lcccc}
    \toprule
    style &day-REL  & day-RMS & night-REL & night-RMS \\ 
    \midrule
    {\ul score}        &   \textbf{0.089}  &     \textbf{3.311}       & 	 \textbf{0.104}      & \textbf{3.831}    \\
    add         &  0.091   &        3.342   &     \textbf{0.104}     & 3.880 \\
    \bottomrule
    \end{tabular}
    }
\end{table}

\begin{table}[t]
\centering
\caption{Update scale Ablations.}
\label{table:ablation_upscale}
\addtolength{\tabcolsep}{1.0pt}
\resizebox{0.45\textwidth}{!}{
    \begin{tabular}{lccccc}
    \toprule
    scale &day-REL  & day-RMS & night-REL & night-RMS & Time (ms) \\ 
    \midrule
    1/4        &   \textbf{0.089}  &     3.376       & 	\textbf{0.101}        &   \textbf{3.789}    & 99 \\
    {\ul 1/8}         &    \textbf{0.089}  &     \textbf{3.311}       & 	 0.104      & 3.831    &  83\\
    1/16        &  0.094   &   3.514        &    0.103  &    3.874  &  82\\
    \bottomrule
    \end{tabular}
    }
\end{table}

\begin{table}[t]
\centering
\caption{Iterations Ablations.}
\label{table:ablation_iteration}
\addtolength{\tabcolsep}{1.0pt}
\resizebox{0.45\textwidth}{!}{
    \begin{tabular}{lccccc}
    \toprule
    $G_1\&G_2$ &day-REL  & day-RMS & night-REL & night-RMS & Time (ms)\\ 
    \midrule
    2        &   0.089  &    3.349        & 	0.105        &    3.860  & 72\\
    {\ul 4}         &     0.089  &     3.311       & 	 0.104      & \textbf{3.831}     & 83\\
    6       &    \textbf{0.087}  &   \textbf{3.272}        &   \textbf{0.102}  &  3.860  &    95 \\
    \bottomrule
    \end{tabular}
    }
\end{table}

\subsection{MVSEC}
Table~\ref{table:mvsec} reports the results on MVSEC. Since we share a similar backbone with the image-based IDisc~\cite{idisc}, we extend IDisc to both modalities by adding an event branch with a same feature extractor as our baseline. Our method shows comparable accuracy on outdoor\_day1 and outperforms other methods on outdoor\_night1 sequence. Compared with the existing state-of-the-art HMNet, PCDepth trained on synthetic Eventscape improves 0.036 in REL and 0.783 in RMS. 

Furthermore, considering the large gap between daytime and nighttime scenarios, we incorporate outdoor\_night2 and outdoor\_night3 into training data to unleash the potential of the proposed pattern-level complementary learning. As a result, PCDepth outperforms HMNet on outdoor\_night1 in $a1$, REL and RMS with a margin of 0.135, 0.078 and 4.103. Besides, compared with our baseline which also introduces event modality at pattern level on outdoor\_night1, our method is still leading, improving 0.148 in REL and 0.003 in RMSlog, which demonstrates the superiority of score-based pattern fusion.

We exhibit qualitative results in Fig.~\ref{fig:qua_mvsec}. The proposed PCDepth can better distinguish objects at different depths (\emph{e.g.}, different trees at row 1, the car and background at row 2 and the building and background at row 4). The improvements are attributed to complementary visual representation learning, which adaptively fuses content from image modality and object contours from event modality.

\subsection{DSEC}
To further verify the superiority of PCDepth, we conduct experiments on DSEC dataset in Table~\ref{table:dsec}. By incorporating event data, all cross-modalities approaches perform better than the original image-based IDisc. Meanwhile, under the same iteration setting, PCDepth (PCDepth-G2) outperforms Baseline-G4 (Baseline-G2) at all 6 metrics across daytime and nighttime scenarios.

Qualitative results are exhibited in Fig.~\ref{fig:qua_dsec}. In challenging regions such as edges of different objects (traffic lights, traffic signs and bus stations), IDisc only takes advantage of image modality and generates low-quality predictions. The baseline method introduces event modalities, improving accuracy at pattern boundaries (bus station at row 4). Our PCDepth exploits the score-based fusion module to integrate patterns across daytime and nighttime scenarios, complementing objects and generating high-quality predictions. Besides, Fig~\ref{fig:attn_maps} visualizes attention maps of scene discretization on DSEC dataset. At each column, two different visual tokens compress different scene objects, thus enabling the network to distinguish objects at different depths.
\subsection{Ablations}
We conduct ablation experiments to validate the proposed improvements. All ablation models are evaluated on DSEC.

\textbf{Complementary visual representation learning.} The complementary visual representation is achieved by the score-based fusion of two sets of visual tokens, which contributes to the adaptation to different circumstances. We ablate the fusion operator by replacing it with a direct add operator in Table~\ref{table:ablation_fuse}. The score-based fusion module achieves better accuracy across daytime and night scenarios.

\textbf{Depth estimator.} Considering the heavy computation for projecting visual patterns back to multiple spatial scales ($1/4, 1/8, 1/16$), we set the depth estimation at 1/8 spatial scale solely. We compare the inference time of the depth estimator and accuracy at different spatial scales in Table~\ref{table:ablation_upscale}. Depth estimation at 1/8 spatial scale showcases a good accuracy-efficiency balance.
 
\textbf{Iterations.} The quality of the complementary visual representation and depth predictions are affected by $G_1$ and $G_2$. We compare different iterations in Table~\ref{table:ablation_iteration}. More iterations lead to better accuracy but lower inference speed, so we set $G_1=G_2=4$ to maintain an accuracy-efficiency trade-off.
\section{CONCLUSION}
In this paper, we propose a novel pattern-based complementary learning architecture for MDE task called PCDepth, which discretizes scenes of image and event modalities into visual tokens and conducts pattern-level fusion for high-quality depth prediction. We build our model based on that existing methods mostly fuse two modalities at pixel level while ignoring that the complementarity mainly impacts high-level patterns which only occupy a few pixels. By exploring the pattern-level complementarity, PCDepth can adaptively integrate images and events and thus boost accuracy, especially in nighttime scenes. Extensive experiments on MVSEC and DSEC demonstrate the superiority of PCDepth. Meanwhile, we notice that pattern-based complementary learning cannot perfectly implement different objects (noise in Fig.~\ref{fig:attn_maps}) and will improve it in future work.

\addtolength{\textheight}{-12cm}   








\bibliographystyle{IEEEtranS}
\bibliography{IEEEabrv, refs}

\begin{thebibliography}{10}
\providecommand{\url}[1]{#1}
\csname url@rmstyle\endcsname
\providecommand{\newblock}{\relax}
\providecommand{\bibinfo}[2]{#2}
\providecommand\BIBentrySTDinterwordspacing{\spaceskip=0pt\relax}
\providecommand\BIBentryALTinterwordstretchfactor{4}
\providecommand\BIBentryALTinterwordspacing{\spaceskip=\fontdimen2\font plus
\BIBentryALTinterwordstretchfactor\fontdimen3\font minus
  \fontdimen4\font\relax}
\providecommand\BIBforeignlanguage[2]{{%
\expandafter\ifx\csname l@#1\endcsname\relax
\typeout{** WARNING: IEEEtran.bst: No hyphenation pattern has been}%
\typeout{** loaded for the language `#1'. Using the pattern for}%
\typeout{** the default language instead.}%
\else
\language=\csname l@#1\endcsname
\fi
#2}}

\bibitem{tore}
R.~W. Baldwin, R.~Liu, M.~Almatrafi, V.~Asari, and K.~Hirakawa, ``Time-ordered
  recent event (tore) volumes for event cameras,'' \emph{IEEE Transactions on
  Pattern Analysis and Machine Intelligence}, vol.~45, no.~2, pp. 2519--2532,
  2022.

\bibitem{adabins}
S.~F. Bhat, I.~Alhashim, and P.~Wonka, ``Adabins: Depth estimation using
  adaptive bins,'' in \emph{Proceedings of the IEEE/CVF Conference on Computer
  Vision and Pattern Recognition}, 2021, pp. 4009--4018.

\bibitem{localbins}
S.~F. Bhat, I.~Alhashim, and P.~Wonka, ``Localbins: Improving depth estimation
  by learning local distributions,'' in \emph{European Conference on Computer
  Vision}.\hskip 1em plus 0.5em minus 0.4em\relax Springer, 2022, pp. 480--496.

\bibitem{surveychen}
G.~Chen, H.~Cao, J.~Conradt, H.~Tang, F.~Rohrbein, and A.~Knoll, ``Event-based
  neuromorphic vision for autonomous driving: A paradigm shift for bio-inspired
  visual sensing and perception,'' \emph{IEEE Signal Processing Magazine},
  vol.~37, no.~4, pp. 34--49, 2020.

\bibitem{transformer}
\BIBentryALTinterwordspacing
A.~Dosovitskiy, L.~Beyer, A.~Kolesnikov, D.~Weissenborn, X.~Zhai,
  T.~Unterthiner, M.~Dehghani, M.~Minderer, G.~Heigold, S.~Gelly, J.~Uszkoreit,
  and N.~Houlsby, ``An image is worth 16x16 words: Transformers for image
  recognition at scale,'' in \emph{International Conference on Learning
  Representations}, 2021. [Online]. Available:
  \url{https://openreview.net/forum?id=YicbFdNTTy}
\BIBentrySTDinterwordspacing

\bibitem{eigen}
D.~Eigen, C.~Puhrsch, and R.~Fergus, ``Depth map prediction from a single image
  using a multi-scale deep network,'' \emph{Advances in neural information
  processing systems}, vol.~27, 2014.

\bibitem{dorn}
H.~Fu, M.~Gong, C.~Wang, K.~Batmanghelich, and D.~Tao, ``Deep ordinal
  regression network for monocular depth estimation,'' in \emph{Proceedings of
  the IEEE conference on computer vision and pattern recognition}, 2018, pp.
  2002--2011.

\bibitem{survey}
G.~Gallego, T.~Delbr{\"u}ck, G.~Orchard, C.~Bartolozzi, B.~Taba, A.~Censi,
  S.~Leutenegger, A.~J. Davison, J.~Conradt, K.~Daniilidis, \emph{et~al.},
  ``Event-based vision: A survey,'' \emph{IEEE transactions on pattern analysis
  and machine intelligence}, vol.~44, no.~1, pp. 154--180, 2020.

\bibitem{learnrep}
D.~Gehrig, A.~Loquercio, K.~G. Derpanis, and D.~Scaramuzza, ``End-to-end
  learning of representations for asynchronous event-based data,'' in
  \emph{Proceedings of the IEEE/CVF International Conference on Computer
  Vision}, 2019, pp. 5633--5643.

\bibitem{ramnet}
D.~Gehrig, M.~R{\"u}egg, M.~Gehrig, J.~Hidalgo-Carri{\'o}, and D.~Scaramuzza,
  ``Combining events and frames using recurrent asynchronous multimodal
  networks for monocular depth prediction,'' \emph{IEEE Robotics and Automation
  Letters}, vol.~6, no.~2, pp. 2822--2829, 2021.

\bibitem{dsec}
M.~Gehrig, W.~Aarents, D.~Gehrig, and D.~Scaramuzza, ``Dsec: A stereo event
  camera dataset for driving scenarios,'' \emph{IEEE Robotics and Automation
  Letters}, vol.~6, no.~3, pp. 4947--4954, 2021.

\bibitem{hmnet}
R.~Hamaguchi, Y.~Furukawa, M.~Onishi, and K.~Sakurada, ``Hierarchical neural
  memory network for low latency event processing,'' in \emph{Proceedings of
  the IEEE/CVF Conference on Computer Vision and Pattern Recognition}, 2023,
  pp. 22\,867--22\,876.

\bibitem{eaodo}
J.~Hidalgo-Carri{\'o}, G.~Gallego, and D.~Scaramuzza, ``Event-aided direct
  sparse odometry,'' in \emph{Proceedings of the IEEE/CVF Conference on
  Computer Vision and Pattern Recognition}, 2022, pp. 5781--5790.

\bibitem{e2depth}
J.~Hidalgo-Carri{\'o}, D.~Gehrig, and D.~Scaramuzza, ``Learning monocular dense
  depth from events,'' in \emph{2020 International Conference on 3D Vision
  (3DV)}.\hskip 1em plus 0.5em minus 0.4em\relax IEEE, 2020, pp. 534--542.

\bibitem{crf1}
P.~Kr{\"a}henb{\"u}hl and V.~Koltun, ``Efficient inference in fully connected
  crfs with gaussian edge potentials,'' \emph{Advances in neural information
  processing systems}, vol.~24, 2011.

\bibitem{fcrn}
I.~Laina, C.~Rupprecht, V.~Belagiannis, F.~Tombari, and N.~Navab, ``Deeper
  depth prediction with fully convolutional residual networks,'' in \emph{2016
  Fourth international conference on 3D vision (3DV)}.\hskip 1em plus 0.5em
  minus 0.4em\relax IEEE, 2016, pp. 239--248.

\bibitem{practicalstereo}
J.~Li, P.~Wang, P.~Xiong, T.~Cai, Z.~Yan, L.~Yang, J.~Liu, H.~Fan, and S.~Liu,
  ``Practical stereo matching via cascaded recurrent network with adaptive
  correlation,'' in \emph{Proceedings of the IEEE/CVF conference on computer
  vision and pattern recognition}, 2022, pp. 16\,263--16\,272.

\bibitem{binsformer}
Z.~Li, X.~Wang, X.~Liu, and J.~Jiang, ``Binsformer: Revisiting adaptive bins
  for monocular depth estimation,'' \emph{arXiv preprint arXiv:2204.00987},
  2022.

\bibitem{erformer}
X.~Liu, J.~Li, X.~Fan, and Y.~Tian, ``Event-based monocular dense depth
  estimation with recurrent transformers,'' \emph{arXiv preprint
  arXiv:2212.02791}, 2022.

\bibitem{slotattention}
F.~Locatello, D.~Weissenborn, T.~Unterthiner, A.~Mahendran, G.~Heigold,
  J.~Uszkoreit, A.~Dosovitskiy, and T.~Kipf, ``Object-centric learning with
  slot attention,'' \emph{Advances in Neural Information Processing Systems},
  vol.~33, pp. 11\,525--11\,538, 2020.

\bibitem{adamw}
I.~Loshchilov and F.~Hutter, ``Fixing weight decay regularization in adam,''
  2017.

\bibitem{srfnet}
T.~Pan, Z.~Cao, and L.~Wang, ``Srfnet: Monocular depth estimation with
  fine-grained structure via spatial reliability-oriented fusion of frames and
  events,'' \emph{arXiv preprint arXiv:2309.12842}, 2023.

\bibitem{idisc}
L.~Piccinelli, C.~Sakaridis, and F.~Yu, ``idisc: Internal discretization for
  monocular depth estimation,'' in \emph{Proceedings of the IEEE/CVF Conference
  on Computer Vision and Pattern Recognition}, 2023, pp. 21\,477--21\,487.

\bibitem{dpt}
R.~Ranftl, A.~Bochkovskiy, and V.~Koltun, ``Vision transformers for dense
  prediction,'' in \emph{Proceedings of the IEEE/CVF international conference
  on computer vision}, 2021, pp. 12\,179--12\,188.

\bibitem{unet}
O.~Ronneberger, P.~Fischer, and T.~Brox, ``U-net: Convolutional networks for
  biomedical image segmentation,'' in \emph{Medical Image Computing and
  Computer-Assisted Intervention--MICCAI 2015: 18th International Conference,
  Munich, Germany, October 5-9, 2015, Proceedings, Part III 18}.\hskip 1em plus
  0.5em minus 0.4em\relax Springer, 2015, pp. 234--241.

\bibitem{iebins}
S.~Shao, Z.~Pei, X.~Wu, Z.~Liu, W.~Chen, and Z.~Li, ``Iebins: Iterative elastic
  bins for monocular depth estimation,'' in \emph{Thirty-seventh Conference on
  Neural Information Processing Systems}, 2023.

\bibitem{ofdepth}
D.~Shi, L.~Jing, R.~Li, Z.~Liu, L.~Wang, H.~Xu, and Y.~Zhang, ``Improved
  event-based dense depth estimation via optical flow compensation,'' in
  \emph{2023 IEEE International Conference on Robotics and Automation
  (ICRA)}.\hskip 1em plus 0.5em minus 0.4em\relax IEEE, 2023, pp. 4902--4908.

\bibitem{even}
P.~Shi, J.~Peng, J.~Qiu, X.~Ju, F.~P.~W. Lo, and B.~Lo, ``Even: An event-based
  framework for monocular depth estimation at adverse night conditions,''
  \emph{arXiv preprint arXiv:2302.03860}, 2023.

\bibitem{pixelshuffle}
W.~Shi, J.~Caballero, F.~Husz{\'a}r, J.~Totz, A.~P. Aitken, R.~Bishop,
  D.~Rueckert, and Z.~Wang, ``Real-time single image and video super-resolution
  using an efficient sub-pixel convolutional neural network,'' in
  \emph{Proceedings of the IEEE conference on computer vision and pattern
  recognition}, 2016, pp. 1874--1883.

\bibitem{onecycle}
L.~N. Smith and N.~Topin, ``Super-convergence: Very fast training of neural
  networks using large learning rates,'' in \emph{Artificial Intelligence and
  Machine Learning for Multi-Domain Operations Applications}, vol. 11006.\hskip
  1em plus 0.5em minus 0.4em\relax International Society for Optics and
  Photonics, 2019, p. 1100612.

\bibitem{raft}
Z.~Teed and J.~Deng, ``Raft: Recurrent all-pairs field transforms for optical
  flow,'' in \emph{Computer Vision--ECCV 2020: 16th European Conference,
  Glasgow, UK, August 23--28, 2020, Proceedings, Part II 16}.\hskip 1em plus
  0.5em minus 0.4em\relax Springer, 2020, pp. 402--419.

\bibitem{stereorep}
S.~Tulyakov, F.~Fleuret, M.~Kiefel, P.~Gehler, and M.~Hirsch, ``Learning an
  event sequence embedding for dense event-based deep stereo,'' in
  \emph{Proceedings of the IEEE/CVF International Conference on Computer
  Vision}, 2019, pp. 1527--1537.

\bibitem{attention}
A.~Vaswani, N.~Shazeer, N.~Parmar, J.~Uszkoreit, L.~Jones, A.~N. Gomez,
  {\L}.~Kaiser, and I.~Polosukhin, ``Attention is all you need,''
  \emph{Advances in neural information processing systems}, vol.~30, 2017.

\bibitem{yeego}
C.~Ye, A.~Mitrokhin, C.~Ferm{\"u}ller, J.~A. Yorke, and Y.~Aloimonos,
  ``Unsupervised learning of dense optical flow, depth and egomotion with
  event-based sensors,'' in \emph{2020 IEEE/RSJ International Conference on
  Intelligent Robots and Systems (IROS)}.\hskip 1em plus 0.5em minus
  0.4em\relax IEEE, 2020, pp. 5831--5838.

\bibitem{crf}
W.~Yuan, X.~Gu, Z.~Dai, S.~Zhu, and P.~Tan, ``Neural window fully-connected
  crfs for monocular depth estimation,'' in \emph{Proceedings of the IEEE/CVF
  Conference on Computer Vision and Pattern Recognition}, 2022, pp. 3916--3925.

\bibitem{surveydeep}
X.~Zheng, Y.~Liu, Y.~Lu, T.~Hua, T.~Pan, W.~Zhang, D.~Tao, and L.~Wang, ``Deep
  learning for event-based vision: A comprehensive survey and benchmarks,''
  \emph{arXiv preprint arXiv:2302.08890}, 2023.

\bibitem{mvsec}
A.~Z. Zhu, D.~Thakur, T.~{\"O}zaslan, B.~Pfrommer, V.~Kumar, and K.~Daniilidis,
  ``The multivehicle stereo event camera dataset: An event camera dataset for
  3d perception,'' \emph{IEEE Robotics and Automation Letters}, vol.~3, no.~3,
  pp. 2032--2039, 2018.

\bibitem{zhuego}
A.~Z. Zhu, L.~Yuan, K.~Chaney, and K.~Daniilidis, ``Unsupervised event-based
  learning of optical flow, depth, and egomotion,'' in \emph{Proceedings of the
  IEEE/CVF Conference on Computer Vision and Pattern Recognition}, 2019, pp.
  989--997.

\end{thebibliography}

\end{document}